\documentclass[10pt,twocolumn,letterpaper]{article}

\usepackage{cvpr}              

\usepackage{bbm}
\usepackage{bm} 
\usepackage{listings}
\usepackage{adjustbox}
\usepackage{booktabs} 
\usepackage{multirow}

\usepackage{pgf}
\usepackage{pgfplots}
\pgfplotsset{compat=1.18}

\usepackage[most]{tcolorbox}
\usepackage{listings}
\usepackage{float}
\usepackage[x11names]{xcolor} 

\usepackage{booktabs}       
\usepackage{amsfonts}       
\usepackage{nicefrac}       
\usepackage{microtype}      
\usepackage{xcolor}         
\usepackage{microtype}
\usepackage{url}
\usepackage{booktabs}
\usepackage{amssymb}
\usepackage[utf8]{inputenc} 
\usepackage[T1]{fontenc}    
\usepackage{url}            
\usepackage{booktabs}       
\usepackage{amsfonts}       
\usepackage{nicefrac}       
\usepackage{microtype}      
\usepackage{xcolor}         
\usepackage{colortbl}
\usepackage{graphicx}
\usepackage{amsmath}
\usepackage{makecell}
\usepackage{multirow}
\usepackage{pifont}
\usepackage[toc,page,header]{appendix}
\usepackage{minitoc}
\usepackage{csquotes}
\usepackage{subcaption}
\usepackage{caption}
\usepackage{adjustbox}
\usepackage{colortbl}
\usepackage[most]{tcolorbox}
\usepackage{lineno}
\usepackage{pgf}
\usepackage{pgfplots}
\usepackage{booktabs}
\usepackage{scalerel}
\usepackage{float}
\usepackage[table,dvipsnames]{xcolor}
\usepackage{booktabs}
\usepackage{caption}
\usepackage{nicematrix} 
\usepackage{tikz} 
\usepackage{xcolor}
\usepackage{soul}             
\usepackage{bm} 
\usepackage{enumitem}
\usepackage{svg}
\usepackage{xspace}
\usepackage{dsfont}

\tcbset{colback=white}
\colorlet{titleblue}{blue!80!black}
\colorlet{titlered}{red!80!black}
\colorlet{titlegreen}{green!80!black}
\colorlet{darkgreen}{green!50!black}

\definecolor{logoRed}{HTML}{CB2F10}
\definecolor{logoBlue}{HTML}{1A4E8A}
\definecolor{logoCyan}{HTML}{56BBCC}

\definecolor{genLLM}{RGB}{250,250,250} 
\definecolor{specLLM}{RGB}{230,230,230}
\definecolor{ourLLM}{RGB}{230,242,230}  
\definecolor{spaLLM}{RGB}{255,249,196}  

\definecolor{genVLM}{RGB}{255,255,255}    
\definecolor{ourVLM}{RGB}{210,210,210} 

\tcbset{
  aibox/.style={
    width=\linewidth,
    top=10pt,
    colback=white,
    colframe=black,
    colbacktitle=black,
    enhanced,
    center,
    attach boxed title to top left={yshift=-0.1in,xshift=0.15in},
    boxed title style={boxrule=0pt,colframe=white,},
  }
}

\newtcolorbox{AIbox}[2][]{aibox,title=#2,#1}

\usepackage{tikz}

\definecolor{cvprblue}{rgb}{0.21,0.49,0.74}
\usepackage[pagebackref,breaklinks,colorlinks,allcolors=cvprblue]{hyperref}

\def\paperID{*****} 
\def\confName{CVPR}
\def\confYear{2026}

\title{MedSAM3: Delving into Segment Anything with Medical Concepts}

\author{Anglin Liu$^{1}$, Xu R. Cao$^{2,\dag}$, Yifan Shen$^{2}$, Yi Lu$^{1}$, Xiang Li$^{2}$, Qianqian Chen$^{3}$, \\Jintai Chen$^{1,4,}$\thanks{Corresponding to: \url{jintaiCHEN@hkust-gz.edu.cn} (J. Chen), \url{xucao2@illinois.edu} (X. Cao).}\\
$^{1}$ The Hong Kong University of Science and Technology (Guangzhou)\\
$^{2}$ University of Illinois Urbana-Champaign\\
$^{3}$ Southeast University\\
$^{4}$ The Hong Kong University of Science and Technology\\
}

\begin{document}
\maketitle

\begin{abstract}
Medical image segmentation is fundamental for biomedical discovery. Existing methods lack generalizability and demand extensive, time-consuming manual annotation for new clinical application. Here, we propose MedSAM-3, a text promptable medical segmentation model for medical image and video segmentation. By fine-tuning the Segment Anything Model (SAM) 3 architecture on medical images paired with semantic conceptual labels, our MedSAM-3 enables medical Promptable Concept Segmentation (PCS), allowing precise targeting of anatomical structures via open-vocabulary text descriptions rather than solely geometric prompts. We further introduce the MedSAM-3 Agent, a framework that integrates Multimodal Large Language Models (MLLMs) to perform complex reasoning and iterative refinement in an agent-in-the-loop workflow. Comprehensive experiments across diverse medical imaging modalities, including X-ray, MRI, Ultrasound, CT, and video, demonstrate that our approach significantly outperforms existing specialist and foundation models. We will release our code and model at \url{https://github.com/Joey-S-Liu/MedSAM3}.
\end{abstract}

\section{Introduction}

\begin{figure*}[t!]
	\centering
	\includegraphics[width=1\textwidth]{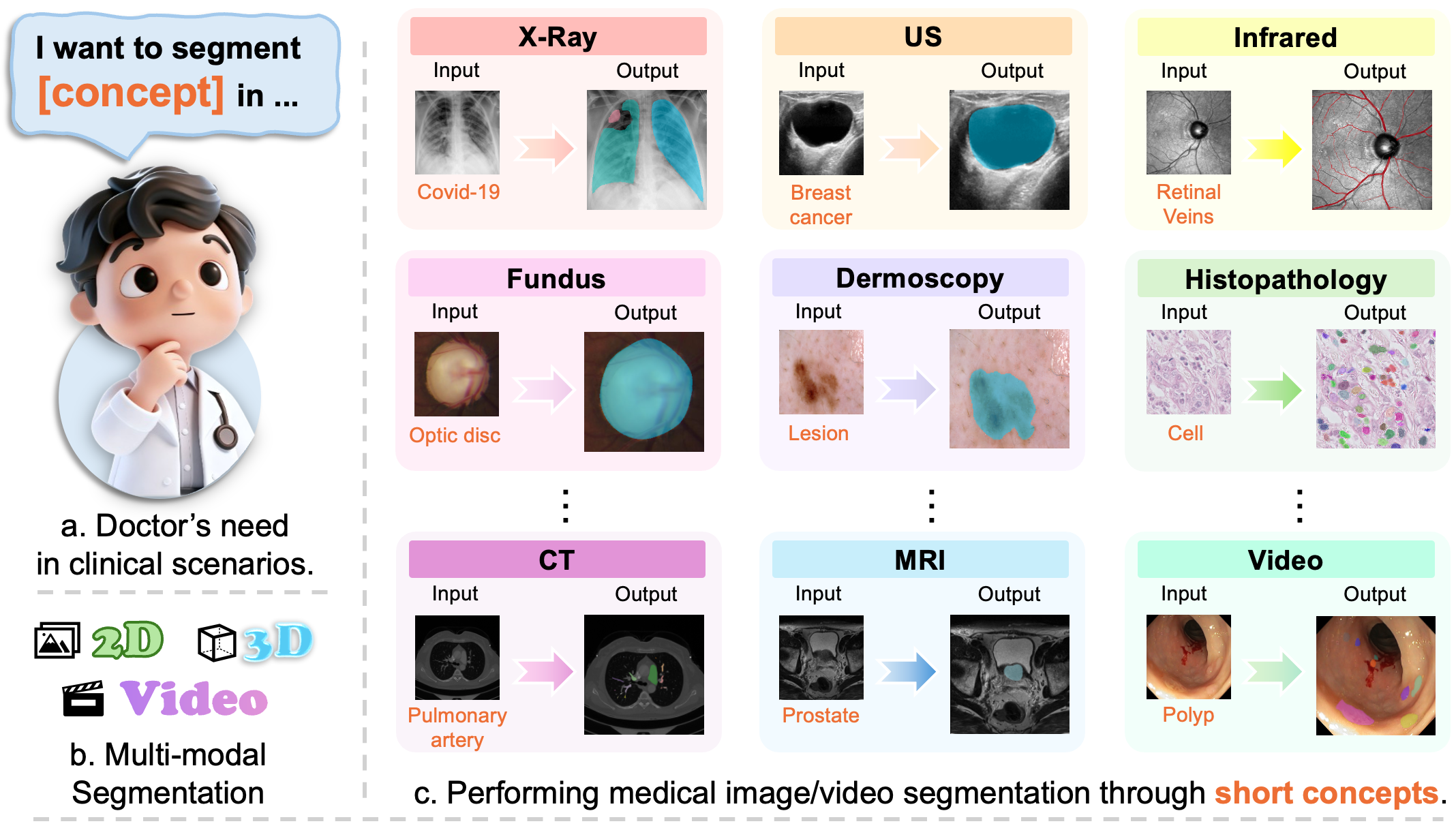}
	\caption{Overview of concept-driven medical image and video segmentation across multiple modalities using MedSAM-3, highlighting that concise clinical concepts directly guide MedSAM-3 to produce reliable segmentations and thereby simplify physicians’ workflow.}
    \label{fig:overview}
    \vspace{-2mm}
\end{figure*}

Medical segmentation is the cornerstone of the modern healthcare system, providing the quantitative analysis necessary for accurate diagnosis, precise treatment planning, and effective monitoring of disease progression~\cite{antonelli2022medical}. While deep learning has driven considerable progress, the development of specialist models for every unique task, modality, and pathology is inefficient and scales poorly. Such models lack generalizability and demand extensive, time-consuming manual annotation for each new clinical application.

The emergence of large-scale foundation models, such as the Segment Anything Model (SAM)~\cite{kirillov2023segment,ravi2024sam}, has marked a paradigm shift towards building generalist systems that can handle diverse tasks. In the medical field, this approach was successfully validated by models like MedSAM~\cite{ma2024segment}, MedSAM-2~\cite{zhu2024medical} and MedSAM2~\cite{ma2025medsam2}, which adapted the original SAM for medical-specific challenges. MedSAM2, in particular, demonstrated the power of a promptable foundation model for segmenting 3D medical images and videos, proving that such systems can drastically reduce manual annotation costs~\cite{ma2025medsam2}. However, these models primarily rely on geometric prompts, which can still be laborious for complex structures and do not fully capture the rich semantic intent of clinicians. In addition, these models can only serve as one tool, lacking potential to connect with the agentic ecosystem supported by multimodal large language models (LLMs)~\cite{li2024mmedagent,yao2025survey,al2025agentic}.

The recent introduction of SAM 3 marks a significant leap in interactive segmentation with its ``Promptable Concept Segmentation'' (PCS) capability~\cite{anonymous2025sam}. Unlike methods reliant on geometric cues, SAM 3 can detect and segment objects based on open-vocabulary conceptual prompts, such as natural language descriptions (e.g., ``a yellow school bus'') or visual exemplars. This ability to operate on semantic concepts presents a transformative opportunity for medical imaging, where clinical language is inherently conceptual (e.g., ``segment the tumor and surrounding edema'' or ``identify all enlarged lymph nodes''). This directly addresses a fundamental limitation of prior text-guidance segmentation models, such as BiomedParse~\cite{zhao2025foundation}, which were constrained to a fixed, pre-defined vocabulary and thus could not generalize to the vast and nuanced range of concepts encountered in clinical practice~\cite{zhao2024biomedparse}.

To address these limitations, we present MedSAM-3, a concept-driven framework designed to segment medical imagery through semantic guidance (Figure~\ref{fig:overview}). We began by benchmarking the original SAM 3 on multiple medical segmentation datasets to validate its baseline capabilities on both text prompting and visual prompting. However, raw SAM 3 struggled with the healthcare domain. Consequently, we fine-tuned the architecture on a curated dataset of diverse medical images paired with rich conceptual labels. The resulting model allows users to segment complex anatomical structures and pathologies using simple text descriptions or visual references from inter- or intra-scan examples. This paradigm shifts the interaction from simple geometric prompting to an intuitive, clinically aligned semantic workflow. Through comprehensive experiments, we demonstrate that MedSAM-3 not only establishes a new state-of-the-art for generalist medical segmentation but also significantly streamlines clinical annotation, paving the way for more intelligent, collaborative medical AI systems.

Our main contributions are:
\begin{itemize} 
\item We propose MedSAM-3, adapting the SAM 3 architecture to the medical domain to enable precise Promptable Concept Segmentation (PCS) using medical text and visual prompts.

\item We introduce the MedSAM3 Agent, an agentic framework that extends MedSAM-3 to process complex, long-form clinical instructions and improve accuracy through an iterative agent-in-the-loop paradigm.

\item We conduct extensive experiments across diverse medical imaging modalities, demonstrating the effectiveness of our design and providing valuable insights into the deployment of concept-based segmentation models in healthcare.
\end{itemize}

\section{Related Works}






\noindent \textbf{Segmentation in Medical Images.}
The field of medical image segmentation has witnessed a remarkable evolution, primarily driven by the transition from convolutional neural networks (CNNs) to Transformer-based architectures. Early advancements were cemented by the U-Net series~\cite{ronneberger2015u,milletari2016v,zhou2018unet++,oktay2018attention,isensee2021nnu,huang2020unet}. With the advent of Vision Transformers, researchers sought to overcome the limited receptive field of CNNs, proposing CNN-Transformer hybrid architecture in medical image segmentation~\cite{chen2021transunet,cao2022swin,hatamizadeh2022unetr}. Despite their success, these specialist models typically require training from scratch for specific organs or modalities, limiting their scalability and generalization across the diverse landscape of clinical tasks~\cite{antonelli2022medical,koleilat2024medclip}. The focus has shifted towards developing large-scale foundation models capable of universal segmentation. The SAM~\cite{kirillov2023segment} demonstrated unprecedented zero-shot generalization in natural images, sparking a wave of adaptations for the medical domain. Initial efforts~\cite{ma2024segment,cheng2023sam,wu2025medical} adapted SAM via fine-tuning or adapter layers to handle medical modalities. This was further extended to 3D volumetric data by models~\cite{du2024segvol,wang2024sam,ma2025medsam2,zhu2024medical}, which leverage temporal or spatial consistency. While these models excel at geometric prompting (points/boxes), they often lack semantic understanding. Recent works like BiomedParse~\cite{zhao2025foundation} and UniverSeg~\cite{butoi2023universeg} have attempted to integrate text guidance. However, as noted in recent studies, these systems are often constrained to fixed vocabularies or lack the reasoning capabilities to interpret complex, open-ended clinical concepts, necessitating a shift towards more agentic architectures~\cite{ma2025medsam2,ravi2024sam}. Meanwhile, specialized vision language model (VLM) designed for medical image segmentation have emerged~\cite{li2023lvit,yang2022lavt}. In the 3D domain, models such as M3D-LaMed~\cite{bai2024m3d} and other promptable frameworks~\cite{liu2023clip,lei2025medlsam,li2024language} have further extended these capabilities. However, due to the suboptimal performance and lack of interactivity in these static segmentation VLMs, developing agent-based systems has become a promising future trend for handling complex clinical scenarios.

\noindent \textbf{Segmentation Agent.}
The integration of Large Language Models (LLMs) with vision systems has given rise to "Segmentation Agents" capable of complex reasoning and interactive understanding. This paradigm moves beyond simple instruction following to "Reasoning Segmentation," where the model must interpret implicit queries (e.g., "segment the reason for the patient's pain"). Pioneering works in the general domain include LISA~\cite{lai2024lisa} and PixelLM~\cite{ren2024pixellm}. This direction was further advanced by multimodal agents~\cite{yuan2024osprey,liu2023gres} and SAM 3~\cite{anonymous2025sam}. These agents differ from static models by maintaining a working memory and iteratively refining predictions based on user feedback, a critical feature for high-stakes decision-making processes in healthcare~\cite{yang2023dawn,moor2023foundation}. In the specialized domain of professional workflows, agentic systems are rapidly transforming how experts interact with data across various verticals. In radiology, agents and models such as MedRAX~\cite{fallahpour2025medrax}, LLaVA-Med~\cite{li2024llava}, and RadFM~\cite{wu2025towards} have been proposed. Beyond healthcare, similar trends are observed in other fields with systems like mDocAgent~\cite{han2025mdocagent}, NovelSeek~\cite{team2025novelseek}, ViperGPT~\cite{suris2023vipergpt}, and ChemLLM~\cite{zhang2024chemllm}. Our work unifies these directions by proposing MedSAM3 Agent, an agent tailored specifically for medical segmentation that combines the reasoning of LLMs with the precise, concept-driven segmentation capabilities of the MedSAM-3 architecture.

\section{Methodology}


\subsection{Enabling Medical Concepts in SAM 3}

Our MedSAM-3 is developed as a generalization of MedSAM-2 and MedSAM2, adopting the unified architecture of SAM 3 to support both the novel Promptable Concept Segmentation (PCS) task and the traditional Promptable Visual Segmentation (PVS) tasks. In the PVS setting, the model accepts diverse visual prompts, such as points, boxes, or masks, to spatially and temporally define individual objects for segmentation. For medical PVS, MedSAM-3 retains the box-based prompting strategy supported by MedSAM2; compared to points and masks, bounding boxes offer a less ambiguous method for specifying clinically useful targets, making them particularly effective for delineating organs and lesions. In addition to PVS, MedSAM-3 introduces support for PCS, allowing the model to target objects using short medical noun phrases, including those with positional adjectives. While the original SAM 3 model is optimized for these concise atomic prompts, MedSAM-3 can handle more complex language queries and reasoning by composing the model with a MLLM within an agentic pipeline.

\begin{figure*}[t!]
	\centering
	\includegraphics[width=1\textwidth]{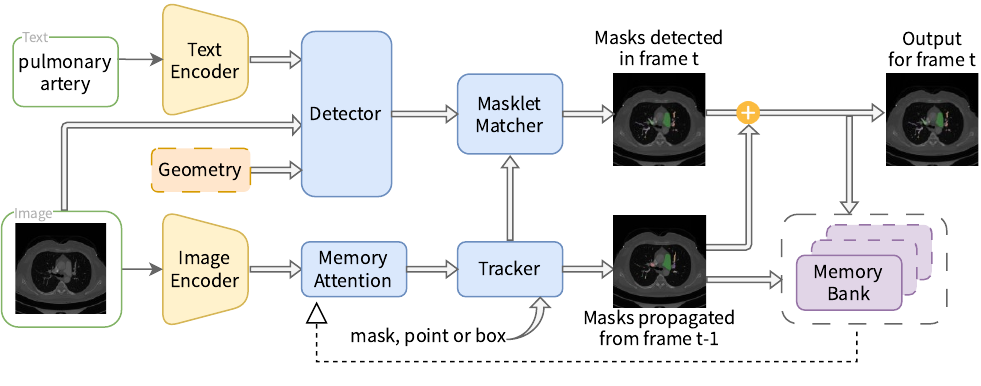}
	\caption{Overview of MedSAM-3.}
    \label{fig:architect}
    \vspace{-3mm}
\end{figure*}

Figure~\ref{fig:architect} illustrates the MedSAM-3 architecture, which features a dual encoder-decoder transformer design. This consists of a detector for image-level capabilities and a tracker paired with a memory module for video tasks. The memory blocks, inherited from SAM 2, employ transformer layers with self-attention and cross-attention mechanisms to condition current frame features on predictions from previous frames via a streaming memory bank. Both the detector and tracker ingest aligned vision-language inputs from a shared Perception Encoder (PE) backbone.

\subsection{Supervised Fine-Tuning with Medical Concepts}
Based on the SAM 3 architecture, the MedSAM-3 model freezes the image and text encoder and updates the remaining detector components during fine-tuning. This design preserves the strong visual and concept prior established by SAM 3 while allowing the task-specific modules to adapt to medical concepts efficiently. The model is optimized using paired medical images and concise concept phrases, each limited to no more than three words and selected strictly according to the dataset’s official documentation or repository descriptions. Such careful curation is motivated by several findings from the SAM 3 evaluation (Section~\ref{sec:discussion}) and is intended to ensure semantic precision, minimize ambiguity in textual guidance, and reduce noise arising from overly broad or inconsistent language. Through this approach, MedSAM-3 strengthens its ability to map semantic medical concepts to anatomically meaningful structures, ultimately enhancing segmentation robustness across diverse clinical scenarios.

\subsection{Scalable Medical Segmentation Agent}

\begin{figure}[t!]
	\centering
	\includegraphics[width=0.43\textwidth]{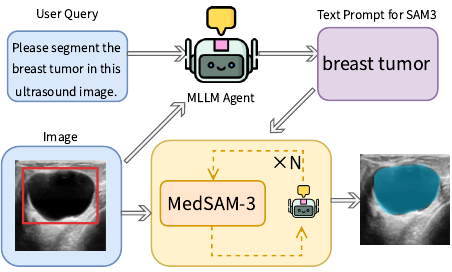}
	\caption{Overview of MedSAM-3 Agent refinement loop. The MedSAM-3 Agent plans and executes multi-step medical image segmentation using a MLLM, generating masks and refining them iteratively with visual and textual feedback.}
    \label{fig:agent}
    \vspace{-2mm}
\end{figure}

We introduce the MedSAM-3 Agent (Figure~\ref{fig:agent}), an agentic framework that dynamically reasons, plans, and executes multistep medical segmentation workflows. Unlike previous approaches, the MedSA-3 Agent integrates multimodal reasoning with concept-guided segmentation capabilities. Given any medical image and a user request, a general MLLM or medical vision-language model (VLM) acts as the core planner: it analyzes the image, devises a step-by-step plan, and invokes MedSAM-3 to generate segmentation masks. After each step, the agent inspects the results, using visual and textual feedback to update its understanding of the environment. This perception-action agent-in-the-loop enables the agent to continuously revise its plan and decide the next action. The process continues until the agent is confident it has satisfied the user's goal (or determines that no valid mask exists), at which point it returns a final set of masks. The resulting pipeline can handle queries far more complex than simple noun phrases, allowing it to understand relationships between anatomical structures and apply visual common sense. We further conduct an experiment to validate a Gemini 3 Pro supported MedSAM-3 Agent can surpass MedSAM-3.

\section{Experiments and Results}
\subsection{Datasets}
To evaluate the segmentation performance of SAM 3 across various medical scenarios and build MedSAM-3, we collected several datasets encompassing multiple imaging modalities, including X-ray, MRI, ultrasound (US), OCT, fundus, dermoscopy, histopathology, nuclear imaging, infrared, endoscopy, and CT, as well as different dimensions, including 2D, 3D, and video. We converted all 3D datasets into frame-sequence formats. When a dataset does not provide an official train–test split, we divide the data using an 80\%–20\% (4:1) ratio. The detailed information of these datasets is presented below.

\begin{itemize}
    \item \textbf{COVID-QU-Ex~\cite{tahir2021covid}.} The COVID-QU-Ex dataset contains 33,920 chest X-ray images and their corresponding infection GTs, including COVID-19 infection, non-COVID-19 infection, and normal cases. A total of 11,956 COVID-19 infection cases were used in this study.
    \item \textbf{BUSI~\cite{al2020dataset}.} The Breast Ultrasound Images Dataset (BUSI) includes 780 breast ultrasound images with corresponding breast cancer GTs, covering standard, benign, and malignant cases.
    \item \textbf{iChallenge-GOALS~\cite{fang2022dataset}.} The Glaucoma OCT Analysis and Layer Segmentation (GOALS) dataset consists of 300 OCT images and their corresponding GTs for the retinal nerve fiber layer, ganglion cell layer, and choroid layer.
    \item \textbf{RIM-ONE~\cite{fumero2011rim}.} The RIM-ONE retinal dataset includes 485 retinal fundus images with corresponding optic disc and cup GTs, comprising 313 images from healthy subjects and 172 from glaucoma patients.
    \item \textbf{ISIC 2018~\cite{codella2019skin}.} The International Skin Imaging Collaboration (ISIC) 2018 dataset contains 2,594 training images and 1,000 test images of skin images, each with associated lesion GTs.
    \item \textbf{MoNuSeg~\cite{kumar2019multi}.} The Multi-organ Nucleus Segmentation Challenge (MoNuSeg) dataset includes 51 histology tissue images from patients with tumors of different organs, each providing a list of annotated nuclei instances.
    \item \textbf{DSB 2018~\cite{caicedo2019nucleus}.} The 2018 Data Science Bowl (DSB 2018) dataset consists of 670 segmented nuclei images acquired under various conditions, differing in cell type, magnification, and imaging modality.
    \item \textbf{RAVIR~\cite{hatamizadeh2022ravir}.} The Retinal Arteries and Veins in Infrared Reflectance Imaging (RAVIR) dataset contains 42 infrared reflectance (IR) images and their corresponding retinal artery and vein GTs.
    \item \textbf{Kvasir-SEG~\cite{jha2019kvasir}.} The Kvasir-SEG dataset includes 1,000 gastrointestinal endoscopy images and their corresponding polyp GTs.
    \item \textbf{Parse2022~\cite{luo2023efficient}.} The Pulmonary Artery Segmentation Challenge 2022(Parse2022) dataset comprises 100 3D lung CT scans with corresponding pulmonary artery GTs.
    \item \textbf{LiTS~\cite{chu2025deep}.} The Liver Tumor Segmentation Challenge (LiTS) dataset contains 130 3D abdominal CT scans and their corresponding liver and liver tumor GTs.
    \item \textbf{PROMISE12~\cite{litjens2014evaluation}.} The Prostate MR Image Segmentation 2012 (PROMISE12) dataset includes 80 3D transversal T2-weighted MRI scans and their corresponding prostate GTs.
    \item \textbf{ISLES 2024~\cite{de2024isles}.} The Ischemic Stroke Lesion Segmentation Challenge 2024 (ISLES 2024) dataset consists of 250 3D brain MRI scans with corresponding ischemic stroke lesion GTs.
    \item \textbf{PolypGen~\cite{ali2023multi}.} The PolypGen dataset is an open-access dataset that comprises 1,537 polyp images, 2,225 positive video sequences with polyp GTs, and 4,275 negative frames.
\end{itemize}

\subsection{Experimental settings}
For the performance evaluation on the 2D datasets, we employed three classical 2D segmentation networks: U-Net~\cite{ronneberger2015u}, Unet3+~\cite{huang2020unet}, and Polyp-PVT~\cite{dong2021polyp}. For the performance evaluation on the 3D datasets, we also adopted three representative 3D segmentation networks: nn-Unet~\cite{isensee2021nnu}, Swin UNETR~\cite{hatamizadeh2021swin}, and U-Mamba~\cite{ma2024u}. To ensure a fair comparison across different methods, all competing approaches except SAM 3 were trained on the training split of each dataset and evaluated on the corresponding test split. SAM 3 was directly tested on the test set without additional training.

For the experiments on the 2D datasets, SAM 3 was evaluated under two settings. In the first setting, the concept input consisted solely of a short phrase describing the target, limited to no more than three words(hereafter referred to as \textbf{SAM 3 T}). In the second setting, the concept input included both the textual phrase and a bounding box enclosing the largest connected component of the target as an image-based reference(hereafter referred to as \textbf{SAM 3 T+I}). For the experiments on the 3D datasets, only the textual phrase was used as the concept input. The phrase inputs used for different datasets are shown in Table~\ref{tab:phrases}. The comprehensive evaluation of SAM 3 is presented in Section~\ref{sec:discussion}.
\begin{table}[t]
  \centering
  {\fontsize{5pt}{6pt}\selectfont
  \resizebox{0.8\linewidth}{!}{%
    \begin{NiceTabular}{l r}[colortbl-like]
      \toprule[1.2pt]
      \textbf{Dataset} & \textbf{Phrase input} \\
      \midrule[1.2pt]
      COVID-QU-Ex      & lung infection \\
      DSB 2018         & nuclei \\
      BUSI             & breast tumor \\
      GOALS            & RNFL \& GCIPL \& choroid \\
      RIM-ONE    & optic cup \& optic disc \\
      ISIC 2018        & skin lesion \\
      RAVIR            & retinal arteries \& retinal veins \\
      Kvasir-SEG       & polyp \\
      MoNuSeg          & nuclei \\
      PolypGen      & polyp \\
      LiTS      & liver \& liver tumor \\
      PROMISE12      & prostate \\
      ISLES 2024      & ischemic stroke lesion \\
      Parse2022      & pulmonary artery \\
      \bottomrule
    \end{NiceTabular}%
  }}
  \caption{The phrase inputs used for different datasets.}
  \label{tab:phrases}
  \vspace{-3mm}
\end{table}

We conducted supervised fine-tuning to build MedSAM-3 on four representative 2D medical datasets from diverse imaging modalities: BUSI, RIM-ONE(Cup), ISIC 2018, and Kvasir-SEG.  In addition, the previous prompt-based SOTA medical segmentation model, MedSAM, was included in the evaluation. Critically, our fine-tuning exclusively targeted the detector module of the underlying SAM 3 architecture, a specific adaptation designed to optimize domain-specific feature detection for medical tasks. To comprehensively assess the model's responsiveness to different types of prompts during the adaptation process, we adopted two distinct fine-tuning paradigms:


\begin{itemize}
    \item 1) Pure Text Prompt Fine-tuning (\textbf{MedSAM-3 T}): In this setting, the model is trained using only the input image and the text description as the prompt. This paradigm aims to enhance the model's ability to ground medical concepts in visual features without explicit spatial guidance.
    \item 2) Text Prompt + Bounding Box Fine-tuning (\textbf{MedSAM-3 T+I}): In this setting, the model is provided with both the semantic text description and a bounding box derived from the ground truth mask. This paradigm evaluates the synergistic effect of combining semantic intent with geometric cues to improve segmentation precision.
\end{itemize}

To evaluate the MedSAM-3 Agent, the test set of the BUSI dataset was used for inference, where the agent model was implemented using Gemini 3 Pro.

All training and inference experiments were conducted on one or two A100 GPUs, each equipped with 80 GB of memory.

\subsection{MedSAM-3 Performance}
\begin{figure}[t!]
	\centering
	\includegraphics[width=0.46\textwidth]{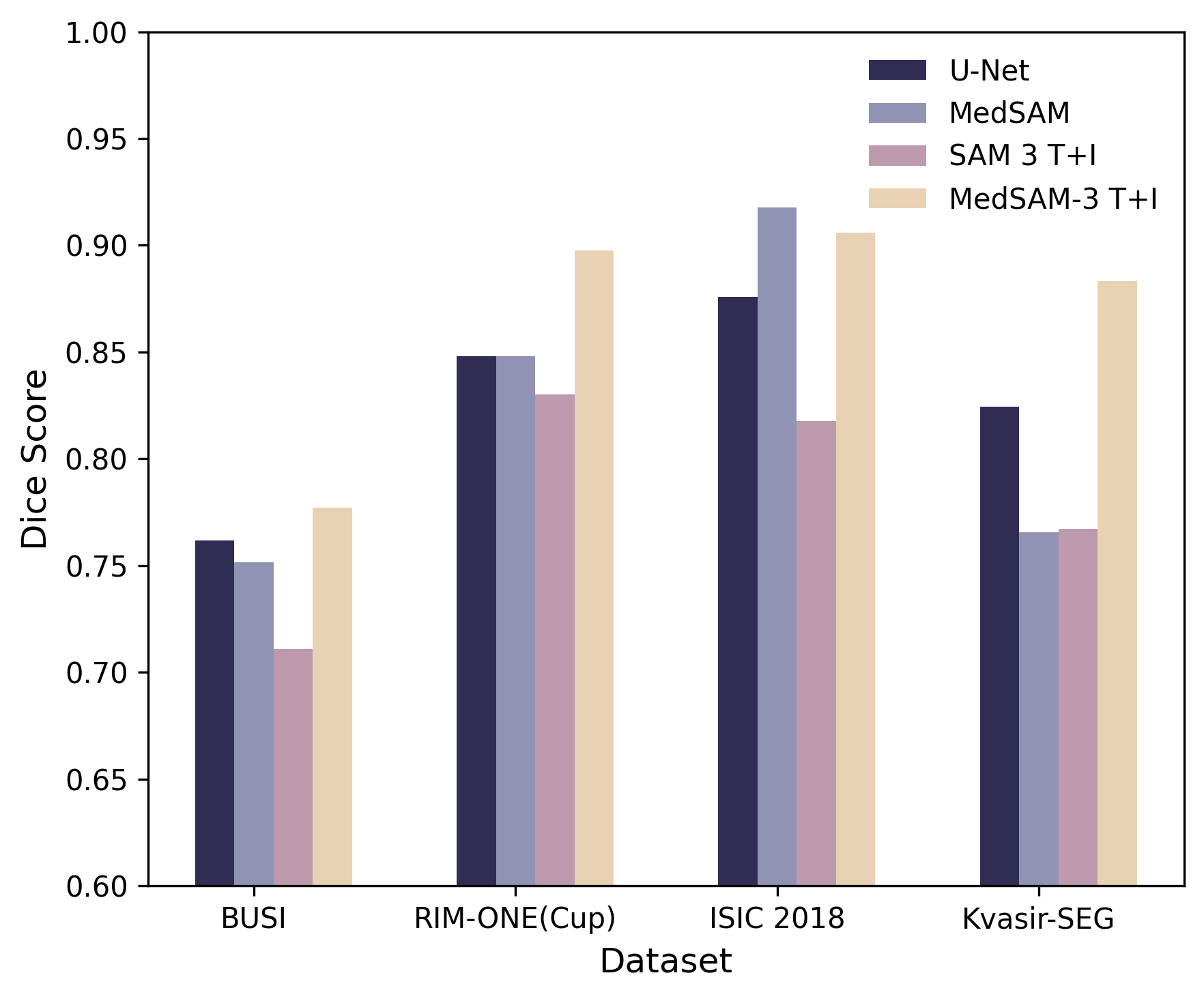}
	\caption{Performance comparison between MedSAM-3 and competing methods on four medical datasets.}
    \label{fig:medsam3_performance}
    \vspace{-2mm}
\end{figure}

The performance of MedSAM-3 is summarized in Table~\ref{tab:MedSAM-3} and Figure~\ref{fig:medsam3_performance}. The text-only inference (MedSAM-3 T) shows clear limitations across all datasets, indicating that text signals alone are insufficient for reliable medical image segmentation. In contrast, the text-and-image inference (MedSAM-3 T+I) yields consistent performance gains and achieves the best results on all four benchmarks, demonstrating visual features with medical-domain text priors enhances robustness across medical imaging conditions. In summary, MedSAM-3 demonstrates strong potential for extension toward universal medical image segmentation.

\begin{figure*}[t!]
	\centering
	\includegraphics[width=0.9\textwidth]{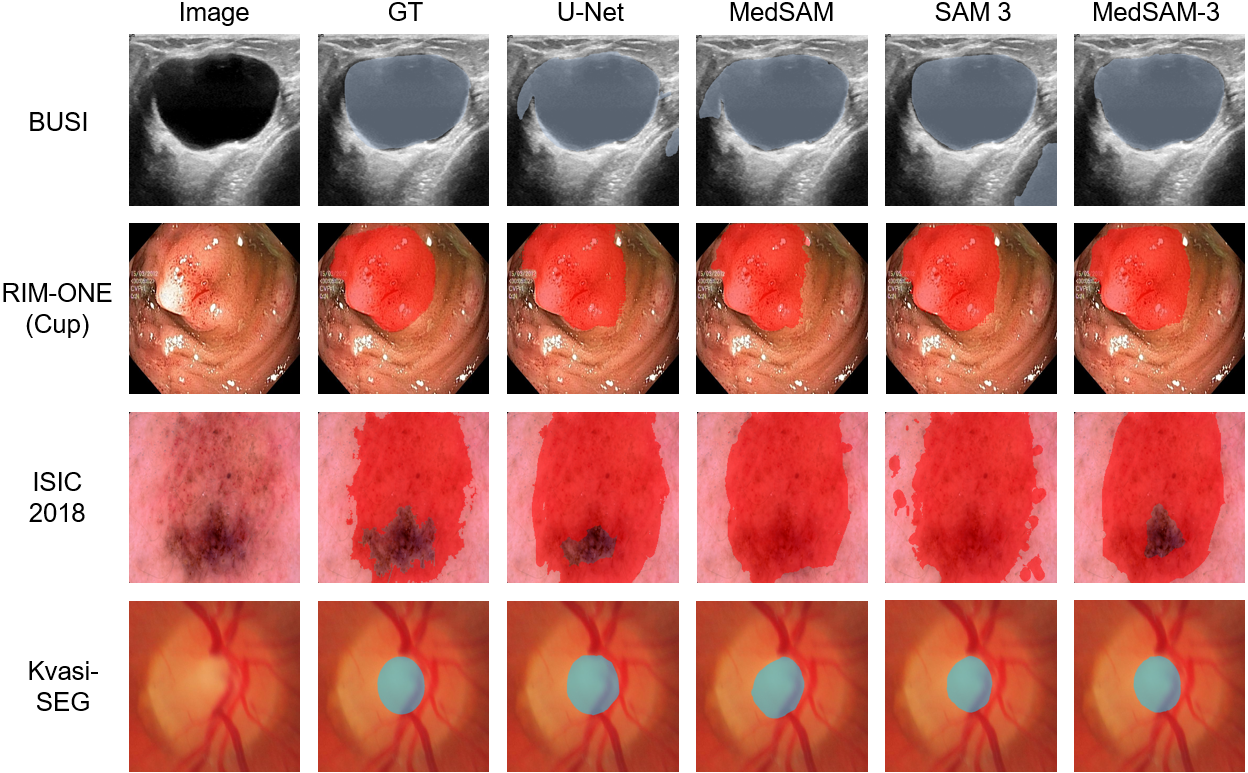}
	\caption{Visualization of the segmentation performance of MedSAM-3, SAM 3(both T+I versions), and other comparison methods.}
    \label{fig:medsam_visual}
    \vspace{-2mm}
\end{figure*}

We also present qualitative visualizations of the segmentation results on the BUSI, RIM-ONE (Cup), ISIC 2018, and Kvasir-SEG datasets. As shown in Figure~\ref{fig:medsam_visual}, MedSAM-3 achieves consistently accurate and visually coherent segmentation across diverse modalities, demonstrating strong performance even in challenging low-contrast or irregular-boundary regions. In contrast, SAM 3 shows noticeable performance degradation. Remarkably, MedSAM-3 attains these improvements with only a small amount of domain-specific fine-tuning data, highlighting the substantial potential of lightweight adaptation for medical image segmentation. More discussion would be posted in Section~\ref{sec:discussion}.

\begin{table}[htbp!]
  \centering
  {\fontsize{9pt}{11pt}\selectfont
  \resizebox{1\linewidth}{!}{%
    \begin{NiceTabular}{l c c c c}[colortbl-like]
      \toprule[1.2pt]
      \textbf{Methods} & \textbf{BUSI} & \textbf{RIM-ONE(Cup)} & \textbf{ISIC 2018} & \textbf{Kvasir-SEG} \\
      \midrule[1.2pt]

      U-Net      & 0.7618 & 0.8480 & 0.8760 & 0.8244 \\
      MedSAM      & 0.7514 & 0.8479 & \textbf{0.9177} & 0.7657 \\
      \rowcolor{genLLM}SAM 3 T      & 0 & 0 & 0.2189 & 0 \\
      \rowcolor{genLLM}SAM 3 T+I    & 0.7110 & 0.8303 & 0.8178 & 0.7671 \\
      \midrule
      \rowcolor{ourVLM}MedSAM-3 T   &
      0.2674 &
      0.0826&
      0.5687&
      0.1441 \\
      \rowcolor{ourVLM}MedSAM-3 T+I   &
      \textbf{0.7772} {\scriptsize\textcolor{green!50!black}{$\uparrow$\,0.0080}} &
      \textbf{0.8977} {\scriptsize\textcolor{green!50!black}{$\uparrow$\,0.0497}} &
      0.9058 {\scriptsize\textcolor{red!70!black}{$\downarrow$\,0.0119}} &
      \textbf{0.8831} {\scriptsize\textcolor{green!50!black}{$\uparrow$\,0.0587}} \\
      
      \bottomrule
    \end{NiceTabular}%
  }}
  \caption{Performance comparison between MedSAM-3 and other methods on four datasets. The best result on each dataset is highlighted in \textbf{bold}. Colored arrows indicate SAM performance changes relative to the best method per dataset.}
  \label{tab:MedSAM-3}
  \vspace{-2mm}
\end{table}

\subsection{MedSAM-3 Agent Performance}
\begin{table}[htbp!]
  \centering
  {\fontsize{9pt}{11pt}\selectfont
  \resizebox{0.75\linewidth}{!}{%
    \begin{NiceTabular}{l c c c}[colortbl-like]
      \toprule[1.2pt]
      \textbf{Methods} & \textbf{MLLM} & \textbf{BUSI}  \\
      \midrule[1.2pt]
      U-Net~\cite{ronneberger2015u}   & -   & 0.7618  \\
      MedSAM~\cite{ma2024segment}   & -   & 0.7514  \\
      MedSAM-3 T+I & -  & 0.7772 \\
      MedSAM-3 Agent & Gemini 3 Pro~\cite{gemini3_2025} & 0.8064  \\
      \bottomrule
    \end{NiceTabular}%
  }}
  \caption{Using Gemini 3 Pro as a controlled and evaluation agent can improve the result.}
  \label{tab:MedSAM3Agent}
  \vspace{-2mm}
\end{table}

We compared the performance of MedSAM-3 against its agentic variant, MedSAM3 Agent, using the BUSI test set (Table~\ref{tab:MedSAM3Agent}). By integrating Gemini 3 Pro~\cite{gemini3_2025} as the core multimodal LLM—orchestrating query interpretation and three rounds of iterative feedback—we achieved a significant performance boost. Specifically, the Dice score improved from 0.7772 to 0.8064. It reveals that agentic workflow could significantly enhance the performance of segmentation models.

\section{Discussion}
\label{sec:discussion}
Prior to developing MedSAM-3, we evaluated the off-the-shelf capabilities of SAM 3 within the medical segmentation domain. Our experiments revealed that the standard SAM 3 model lacks the necessary generalization to handle diverse clinical modalities effectively. These limitations motivated the design of MedSAM-3 and the agentic MedSAM-3 Agent framework.
\subsection{SAM 3 Performance}
\subsubsection{SAM 3 Performance on 2D/Video Datasets}
\begin{figure}[t!]
	\centering
	\includegraphics[width=0.5\textwidth]{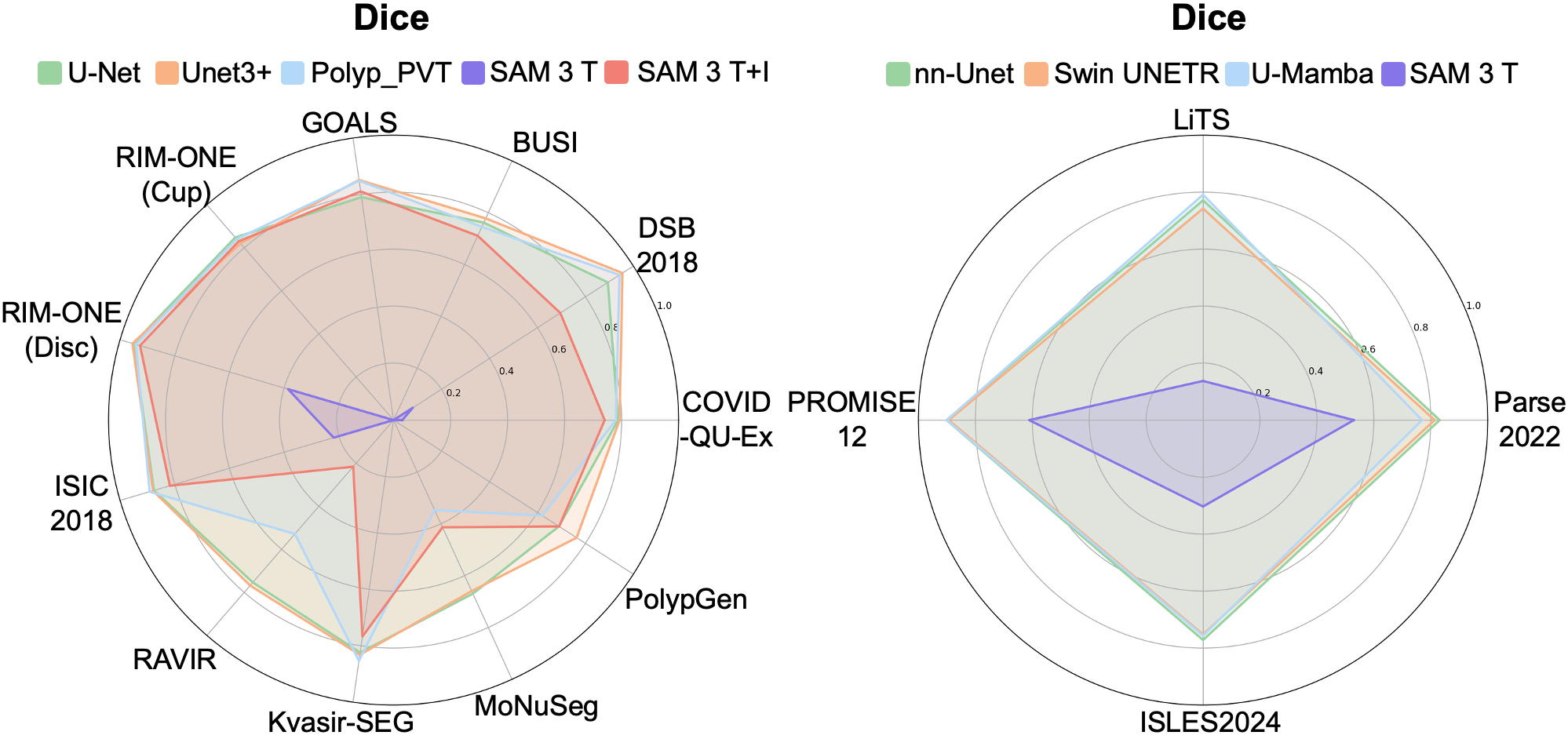}
	\caption{Radar charts of different models’ performance on different datasets. Left: 2D/video scene; Right: 3D scene.}
    \label{fig:radar}
    \vspace{-2mm}
\end{figure}

Table \ref{tab:main_experiment} summarizes the segmentation results of SAM 3 on several representative 2D/video medical imaging datasets. Overall, SAM 3 exhibits highly uneven performance, showing a strong “subject bias.” In some datasets, such as RIM-ONE, the model achieves relatively high accuracy. However, in datasets like DSB 2018 and RAVIR, it almost completely fails. This inconsistency may arise from the model’s limited sensitivity to medical concepts and its instability in distinguishing semantically similar medical terms. Furthermore, even though part of the evaluation data may have been included in SAM 3’s pretraining corpus, its transferability to medical scenarios remains weak, revealing the difficulty of cross-domain generalization.

Notably, incorporating bounding box guidance (Text + BBX) leads to a substantial improvement in segmentation quality, where the Dice scores approach or even surpass those of conventional supervised methods. This demonstrates the critical role of geometric cues in assisting conceptual understanding, consistent with the findings in the natural domain. In summary, SAM 3 performs inconsistently across 2D medical segmentation tasks, with strong reliance on spatial hints to compensate for its limited grasp of fine-grained medical semantics. This observation suggests that combining conceptual and geometric information remains essential for achieving reliable segmentation in medical imaging.

\begin{table*}[htbp!]
  \centering
  {\fontsize{9pt}{11pt}\selectfont
  \resizebox{0.99\linewidth}{!}{%
    \begin{NiceTabular}{l c c c c c c c c c c c}[colortbl-like]
      \toprule[1.2pt]
      \textbf{Methods} &
      \textbf{COVID-QU-Ex} & \textbf{DSB 2018} & \textbf{BUSI} & \textbf{GOALS} &
      \textbf{RIM-ONE(Cup)} & \textbf{RIM-ONE(Disc)} & \textbf{ISIC 2018} &
      \textbf{RAVIR} & \textbf{Kvasir-SEG} & \textbf{MoNuSeg} & \textbf{PolypGen} \\ 
      \midrule[1.2pt]

      U-Net &
      0.7880 & 0.8936 & 0.7618 & 0.7902 & \textbf{0.8480} & 0.9514 & 0.8760 & 0.7539 & 0.8244 & \textbf{0.6696} & 0.6897 \\

      Unet3+ &
      \textbf{0.7928} & \textbf{0.9545} & \textbf{0.7782} & \textbf{0.8513} & 0.8206 &
      \textbf{0.9545} & 0.8797 & \textbf{0.7681} & 0.8321 & 0.6595 & \textbf{0.7634} \\

      Polyp-PVT &
      0.7800 & 0.9420 & 0.7457 & 0.8487 & 0.8406 & 0.9420 & \textbf{0.8917} &
      0.5284 & \textbf{0.8536} & 0.3472 & 0.6205 \\
      \midrule

       \rowcolor{ourVLM}SAM 3 T &
      0.0305 {\scriptsize\textcolor{red!70!black}{$\downarrow$\,0.7623}} &
      0.0803 {\scriptsize\textcolor{red!70!black}{$\downarrow$\,0.8742}} &
      0 {\scriptsize\textcolor{red!70!black}{$\downarrow$\,0.7782}} &
      0 {\scriptsize\textcolor{red!70!black}{$\downarrow$\,0.8513}} &
      0 {\scriptsize\textcolor{red!70!black}{$\downarrow$\,0.8480}} &
      0.3858 {\scriptsize\textcolor{red!70!black}{$\downarrow$\,0.5687}} &
      0.2189 {\scriptsize\textcolor{red!70!black}{$\downarrow$\,0.6728}} &
      0 {\scriptsize\textcolor{red!70!black}{$\downarrow$\,0.7681}} &
      0 {\scriptsize\textcolor{red!70!black}{$\downarrow$\,0.8536}} &
      0 {\scriptsize\textcolor{red!70!black}{$\downarrow$\,0.6696}} &
      0 {\scriptsize\textcolor{red!70!black}{$\downarrow$\,0.7634}} \\

       \rowcolor{ourVLM}SAM 3 T + I &
      0.7405 {\scriptsize\textcolor{red!70!black}{$\downarrow$\,0.0523}} &
      0.6953 {\scriptsize\textcolor{red!70!black}{$\downarrow$\,0.2592}} &
      0.7110 {\scriptsize\textcolor{red!70!black}{$\downarrow$\,0.0672}} &
      0.8108 {\scriptsize\textcolor{red!70!black}{$\downarrow$\,0.0405}} &
      0.8303 {\scriptsize\textcolor{red!70!black}{$\downarrow$\,0.0177}} &
      0.9270 {\scriptsize\textcolor{red!70!black}{$\downarrow$\,0.0275}} &
      0.8178 {\scriptsize\textcolor{red!70!black}{$\downarrow$\,0.0739}} &
      0.2163 {\scriptsize\textcolor{red!70!black}{$\downarrow$\,0.5518}} &
      0.7671 {\scriptsize\textcolor{red!70!black}{$\downarrow$\,0.0865}} &
      0.4135 {\scriptsize\textcolor{red!70!black}{$\downarrow$\,0.2561}} &
      0.6903 {\scriptsize\textcolor{red!70!black}{$\downarrow$\,0.0731}} \\

      \bottomrule
    \end{NiceTabular}%
  }}
  \caption{Performance comparison between SAM 3 and traditional segmentation models. The best result on each dataset is highlighted in \textbf{bold}. Colored arrows indicate SAM performance changes relative to the best method per dataset.}
  \label{tab:main_experiment}
  \vspace{-2mm}
\end{table*}

\subsubsection{SAM 3 Performance on 3D Datasets}
Table \ref{tab:main_experiment_3d} summarizes the performance of different methods on several 3D medical image datasets. Overall, nn-UNet, Swin UNETR, and U-Mamba achieve relatively stable and high Dice scores across tasks, whereas SAM 3 shows consistently lower performance on all four datasets, with particularly large gaps on more challenging data such as LiTS and ISLES 2024. These observations indicate that SAM 3 remains limited when applied to volumetric segmentation.

\begin{table}[htbp!]
  \centering
  {\fontsize{9pt}{11pt}\selectfont
  \resizebox{1\linewidth}{!}{%
    \begin{NiceTabular}{l c c c c}[colortbl-like]
      \toprule[1.2pt]
      \textbf{Methods} & \textbf{Parse2022} & \textbf{LiTS} & \textbf{PROMISE12} & \textbf{ISLES2024} \\
      \midrule[1.2pt]

      nn-Unet      & \textbf{0.8311} & 0.7714 & \textbf{0.9011} & \textbf{0.7718} \\
      Swin UNETR   & 0.8134 & 0.7425 & 0.8934 & 0.7523 \\
      U-Mamba      & 0.7692 & \textbf{0.7910} & 0.9002 & 0.7566 \\
      \midrule
      \rowcolor{ourVLM}SAM 3 T   &
      0.5295 {\scriptsize\textcolor{red!70!black}{$\downarrow$\,0.3016}} &
      0.1374 {\scriptsize\textcolor{red!70!black}{$\downarrow$\,0.6536}} &
      0.6110 {\scriptsize\textcolor{red!70!black}{$\downarrow$\,0.2901}} &
      0.3033 {\scriptsize\textcolor{red!70!black}{$\downarrow$\,0.4685}} \\
      
      \bottomrule
    \end{NiceTabular}%
  }}
  \caption{Performance comparison between SAM 3 and other methods on 3D medical image datasets. The best result on each dataset is highlighted in \textbf{bold}. Colored arrows indicate SAM performance changes relative to the best method per dataset.}
  \label{tab:main_experiment_3d}
  \vspace{-2mm}
\end{table}

Figure~\ref{fig:radar} also visually illustrates the substantial performance disparities of SAM 3 across different datasets.
\subsection{Findings}
We summarizes the key observations from applying SAM 3 to medical image segmentation tasks, further fine-tuning the model on domain-specific datasets to develop MedSAM-3, and employing an advanced MLLM to construct the MedSAM-3 Agent. Through a systematic evaluation across different modalities and targets, several recurring patterns were identified. These observations reflect the limitations of SAM 3 when applied directly to medical scenarios and the behaviors that persist or newly emerge after task-specific adaptation. The analysis further indicates that fine-tuning can substantially improve performance, yet its effectiveness is constrained by the limited scale of medical datasets and the scarcity of high-quality data containing rich clinical terminology and domain-specific textual descriptions. Moreover, integrating an MLLM-based agent reveals additional potential of MedSAM-3, enabling more flexible interaction and better utilization of medical knowledge. Together, these findings highlight both the challenges and opportunities in adapting general-purpose vision-language models to meet the precision and structural requirements of medical image analysis. The main findings are outlined below.

\begin{figure*}[htbp!]
	\centering
	\includegraphics[width=1\textwidth]{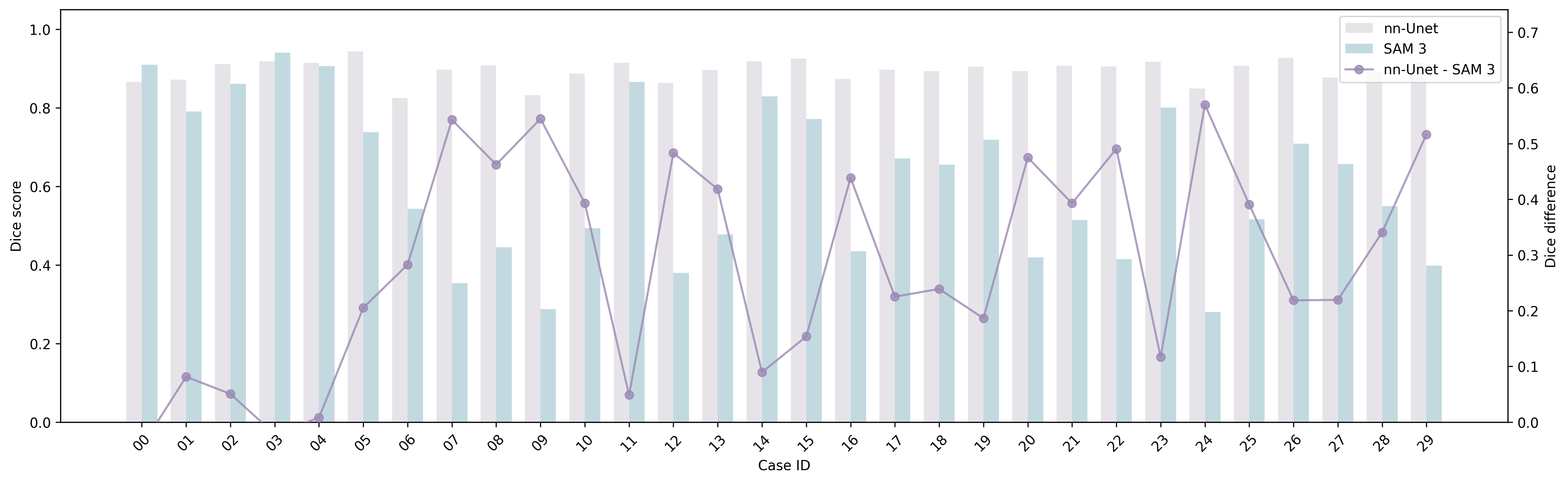}
	\caption{Comparison of per-case performance between SAM 3 and nn-Unet on the PROMISE12 test set. The bar chart shows the absolute Dice values of different methods across cases, while the line plot illustrates the performance differences between the two methods for each case.}
    \label{fig:promise12_gap}
    \vspace{-2mm}
\end{figure*}

\noindent \textbf{Finding 1: Substantial Performance Discrepancy Between SAM 3 and Established Medical Segmentation Baselines.} Across all evaluated datasets, SAM 3 shows a large and unusual performance gap compared with standard medical segmentation models. This pattern is consistent across 2D, video, and 3D tasks. A representative example is the PROMISE12 dataset. Although PROMISE12 has clear anatomy and minimal semantic ambiguity, some are segmented reasonably well while many others fail severely among the 30 test cases shown in Figure~\ref{fig:promise12_gap}.

\begin{figure*}[htbp!]
	\centering
	\includegraphics[width=0.75\textwidth]{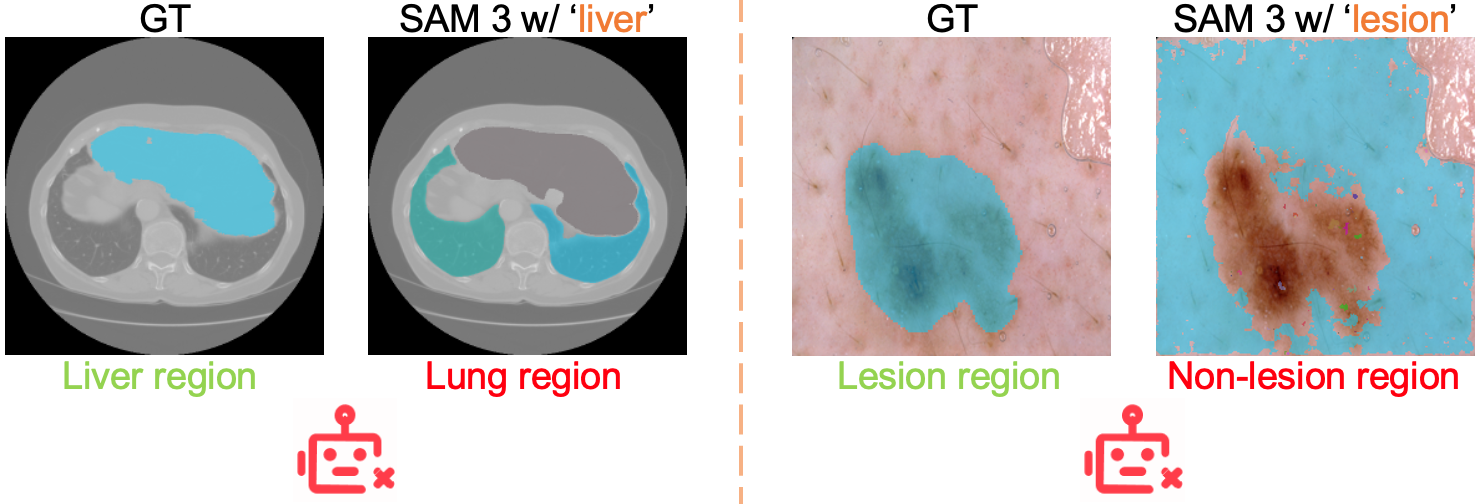}
	\caption{Examples of SAM 3 inference. On the left, using the LiTS dataset, the model segments the lung regions instead of the liver with the concept ``liver''. On the right, using the ISIC2018 dataset, the model segments surrounding non-lesion regions instead of the lesion with the concept ``lesion''.}
    \label{fig:/liver_to_lung}
    \vspace{-2mm}
\end{figure*}

\noindent \textbf{Finding 2: Systematic Misalignment Between Concept Prompts and Anatomical Target Regions in SAM 3.} SAM 3 exhibits a consistent pattern of misalignment between concepts and the regions it predicts, as shown in Figure~\ref{fig:/liver_to_lung}. In the LiTS liver segmentation task, concept specifying “liver” often leads the model to segment the left and right lungs while the true liver region is almost entirely ignored. A similar phenomenon occurs in the ISIC 2018 dataset. When the concept is “lesion”, the model frequently produces a prediction that covers broad non-lesion areas and excludes the actual lesion. The recurrence of this behavior across tasks suggests that SAM 3 does not reliably associate textual prompts with their corresponding visual targets.

\begin{figure}[htbp!]
	\centering
	\includegraphics[width=1.0\linewidth]{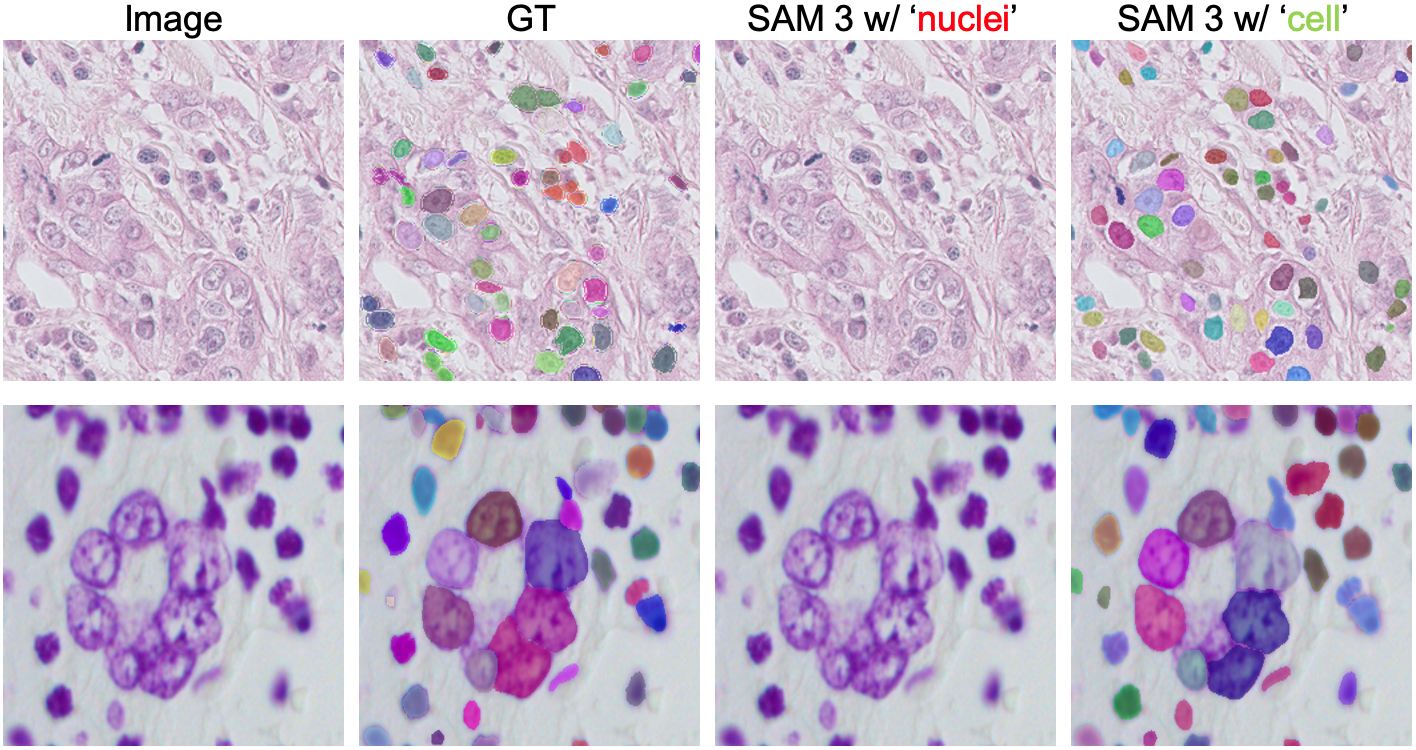}
	\caption{An example illustrating the large performance gap of SAM 3 on the MoNuSeg and DSB2018 dataset when provided with semantically similar concept inputs.}
    \label{fig:/MoNuSeg_example}
    \vspace{-2mm}
\end{figure}

\noindent \textbf{Finding 3: Limited Semantic Discrimination of Fine-Grained Medical Terminology by SAM 3.} The SAM 3 model struggles to distinguish between closely related biological or anatomical concepts. We take the MoNuSeg dataset and DSB 2018 dataset for example, shown in Figure~\ref{fig:/MoNuSeg_example}. The concept ``nucleus'' or ``nuclei'' was totally unrecognized, even the full name of the MoNuSeg dataset is actually the Multi-organ \textbf{Nucleus} Segmentation Challenge, whereas the more generic prompt ``cell'' produced reasonable segmentation results. The details are shown in Table~\ref{tab:monuseg_prompt}.

\begin{table}[htbp!]
  \centering
  {\fontsize{9pt}{12pt}\selectfont
  \resizebox{0.9\linewidth}{!}{%
    \begin{NiceTabular}{l @{\hspace{15mm}} c @{\hspace{15mm}} c c @{\hspace{15mm}} c}[colortbl-like]
      \toprule[1.2pt]
      \Block{2-1}{\textbf{Methods}} & \multicolumn{2}{c}{\textbf{MoNuSeg}} & \multicolumn{2}{c}{\textbf{DSB 2018}} \\ \cline{2-5}
       & \textbf{nuclei} & \textbf{cell} & \textbf{nuclei} & \textbf{cell} \\
      \midrule[1.2pt]

      SAM 3 T     & 0 & 0.6786 &0.0803 &0.5753\\
      SAM 3 T + I & 0.4135 & 0.7907 & 0.6953& 0.7384\\

      \bottomrule
    \end{NiceTabular}%
  }}%
  \caption{Performance of SAM 3 on the MoNuSeg and DSB 2018 under different concepts.}
  \label{tab:monuseg_prompt}
  \vspace{-2mm}
\end{table}

\noindent \textbf{Finding 4: MedSAM-3 Demonstrates Promising Performance Through Efficient Domain Adaptation.} Through domain-specific fine-tuning with curated medical concept annotations, MedSAM-3 shows encouraging improvements in concept alignment and segmentation reliability across diverse clinical imaging modalities. While further scaling would benefit from broader concept-annotated datasets, the current approach demonstrates meaningful progress toward more generalizable medical segmentation systems.

\noindent \textbf{Finding 5: The Agentic Framework Effectively Raises the Performance Ceiling of MedSAM-3.} 
Integrating an MLLM-based agent brings measurable improvements to MedSAM-3, particularly in handling complex clinical instructions and performing iterative refinement. By orchestrating query interpretation through Gemini 3 Pro with iterative feedback loops, the agentic framework elevates segmentation accuracy through multi-step reasoning, intelligent prompt refinement, and error correction. These results highlight the promising potential of agentic workflows to enhance foundation models for increasingly complex and nuanced medical segmentation scenarios in real-world clinical practice.

\subsection{Why MedSAM-3?}
Directly applying SAM 3 in medical scenarios leads to substantial performance degradation because the model lacks the domain-specific semantic grounding required for precise concept–region alignment. Medical structures often exhibit low contrast, subtle boundaries, and significant inter-patient variability, and these properties are not represented in the natural-image corpus used for SAM 3 pre-training. As a result, the model struggles to interpret fine-grained anatomical or pathological terms and frequently produces unstable masks under text-only or weak prompt settings.

MedSAM-3 addresses these limitations by introducing medical concepts with explicit, high-quality supervision. The fine-tuning stage exposes the model to medically meaningful terminology, consistent labeling rules, and clinically relevant spatial patterns, enabling it to learn the specialized semantic relationships absent from the original pre-training. This adaptation substantially improves its reliability across modalities and tasks, allowing the model to generate semantically grounded predictions even when visual cues are subtle or ambiguous.

Our empirical results show that the improvements brought by the agentic framework depend critically on the quality of the underlying model. The agent can refine prompts and perform iterative correction, but its effectiveness diminishes when the base segmentation is semantically misaligned. In contrast, MedSAM-3 provides a stable and medically aligned starting point, ensuring that subsequent agentic reasoning operates on a meaningful foundation.

Overall, MedSAM-3 serves as the essential bridge between general-purpose foundation models and the precision required in clinical image analysis. It transforms SAM 3 from a broadly capable vision-language model into one that can consistently operate within the constraints and expectations of the medical domain, thereby establishing a reliable performance baseline for both direct inference and agent-assisted segmentation.

\section{Conclusion}
We propose MedSAM-3, extending the SAM 3 architecture to address the unique challenges of medical concept grounding. Through domain-specific fine-tuning on PCS task, MedSAM-3 significantly outperforms the original SAM 3, particularly in handling complex medical semantics and temporal consistency. We further enhanced this backbone with the MedSAM-3 Agent, an agent-in-the-loop framework that improves usability via iterative feedback. Our analysis reveals that MedSAM-3 determines the fundamental segmentation quality, while the agent leverages reasoning to correct errors and optimize prompts, effectively pushing the performance ceiling. This work demonstrates the potential of coupling domain adaptation with agentic workflows. Future work will address current limitations in concept granularity and text–image alignment, with the aim of scaling MedSAM-3 to a wider range of clinical applications. We will release our code and models to support the community.

{
    \small
    \bibliographystyle{ieeenat_fullname}
    \bibliography{main}
}


\end{document}